\documentclass[runningheads]{llncs}

\usepackage[T1]{fontenc}
\usepackage{amsmath}
\usepackage{amssymb}
\usepackage{multirow}
\usepackage{booktabs}
\usepackage{xcolor}
\usepackage[most]{tcolorbox}
\usepackage{enumitem}
\usepackage{listings}
\usepackage{courier}
\usepackage{makecell}  % 用于在单元格里换行
\usepackage{diagbox}
\usepackage{marvosym}
\usepackage{varwidth}
\usepackage{graphicx}
\usepackage{hyperref}
\usepackage{color}

\begin{document}
\title{Reinforcement Learning Enhanced LLM Agents for Complex Vehicle Routing Problems}
\titlerunning{RLEA}
\author{Yi Chen\inst{1}\orcidID{0009-0009-6748-0479} \and
Zikang Yu\inst{1}\orcidID{0000-0001-6992-2829} \and
Jiahai Wang\inst{1}${(\textrm{\Letter})}$\orcidID{0000-0002-6961-7813} \and
Jinbiao Chen\inst{2}\orcidID{0000-0001-7417-0430} \and
Jianpeng Zhou\inst{1}\orcidID{0009-0002-4570-3298} \and
Zizhen Zhang\inst{1}\orcidID{0000-0003-0320-9355}}
\authorrunning{Chen. Author et al.}
\institute{School of Computer Science and Engineering, Sun Yat-sen University, Guangzhou 510006, China\\
\email{\{cheny2596,wangjiah,yuzk6,zhoujp7\}@mail2.sysu.edu.cn}\\
\email{\{wangjiah, zhangzzh7\}@mail.sysu.edu.cn}
% \url{http://www.springer.com/gp/computer-science/lncs}
\and
Department of Industrial Systems Engineering and Management, National University of Singapore, Singapore
\\
\email{bill.cjb@nus.edu.sg}}
\maketitle              % typeset the header of the contribution
\begin{abstract}
% Vehicle Routing Problems are fundamental combinatorial optimization challenges with prevalent applications across various scenarios. The state-of-the-art solvers can effectively solve such problems. However, the necessity for expert-level modeling and coding effectively hinders the democratized use of these optimization technologies. 
% This paper presents Reinforcement Learning Enhanced LLM Agents (RLEA) for the automated modeling of complex Vehicle Routing Problems. RLEA includes a lightweight Planner optimized via Soft Q-Learning, which enables rapid, real-time action orchestration and high exploration efficiency. Furthermore, we implement an LLM-based agent augmented by an evolutionary memory module and Retrieval-Augmented Generation (RAG). This integration enables the agent to leverage both internalized experiential insights and extended external knowledge for more robust problem-solving. 
% We conducted a comprehensive evaluation across 48 distinct VRP variants. Experimental results demonstrate that our approach attains a 62.50\% success rate, significantly outperforming current state-of-the-art (SOTA) methods by a margin of 16.67\%. These findings validate the effectiveness of our framework in balancing computational efficiency with robust optimization performance.

Vehicle Routing Problems (VRPs) are fundamental combinatorial optimization problems with widespread applications in various scenarios. The advanced optimization solvers can effectively solve such problems. However, modeling complex VRP variants for solvers often requires substantial domain expertise, which limits the accessibility of advanced optimization technologies.
In this paper, we propose Reinforcement Learning Enhanced LLM Agents (RLEA), a multi-agent framework designed to automate the modeling of complex VRPs. 
RLEA introduces a lightweight neural Planner trained with Soft Q-learning to efficiently orchestrate the actions of LLM-based agents. In addition, we equip the system with an evolutionary memory module and retrieval-augmented generation, enabling the agent to leverage both accumulated experience and external solver knowledge during program generation and refinement for solving VRPs.
We evaluated 48 distinct VRP variants across various solvers. The experimental results demonstrate that RLEA outperforms the previous state-of-the-art method, achieving a 16.67\% higher success rate while significantly reducing runtime errors. These results validate that integrating reinforcement learning with LLM-based reasoning is highly effective for automated optimization modeling. The appendix is available at:~\href{https://doi.org/10.5281/zenodo.19134435}{\textcolor{blue}{https://doi.org/10.5281/zenodo.19134435}}.

\end{abstract}
\section{Introduction}
Vehicle routing problems (VRPs) constitute an important class of combinatorial optimization problems in operations research and have been widely applied across various domains, such as communication and transportation~\cite{wu2024neural,bi2022learning,liao2025bopo}. Recent studies increasingly focus on more complex VRP variants to better reflect real-world scenarios~\cite{mor2022vehicle,yzk}. As the number and complexity of the constraints increase, the challenge of solving these NP-hard VRPs grows commensurately~\cite{chen2022deep}. 

Recently, large language models (LLMs) have been increasingly applied to solving complex VRPs due to their strong capabilities in reasoning and code generation~\cite{feng2026monitor}. The LLM-based methods for complex VRPs can be mainly divided into three categories. 

The first paradigm attempts to directly generate solutions using LLM~\cite{DBLP:journals/corr/abs-2509-16865}. These methods prompt the model to infer routes or decision sequences from natural language descriptions. However, due to the complex combinatorial structure and strict feasibility constraints of VRPs, these methods often struggle to produce valid and scalable solutions.
The second paradigm emphasizes automated heuristic design. Leveraging the reasoning and code generation capabilities of LLMs, this approach has demonstrated substantial potential for automating heuristic design. Recent works~\cite{DBLP:journals/nature/RomeraParedesBNBKDREWFKF24,DBLP:conf/icml/0044TY0LWL024,DBLP:conf/nips/Ye0CBHKPS24,DBLP:conf/icml/0009X0H25,DBLP:conf/ppsn/Cazenave24,DBLP:conf/acl/GuoYKY25,DBLP:conf/ppsn/CybulaJPRS22,DBLP:journals/corr/abs-2510-16701,DBLP:journals/corr/abs-2502-15359} focus on combining evolutionary computation (EC) with LLMs to generate heuristics to solve VRPs. Although these methods show promise, they remain heavily reliant on the internal knowledge of LLMs and fail to leverage the capabilities of state-of-the-art (SOTA) optimization solvers designed by experts.

The third paradigm focuses on automatic modeling~\cite{DBLP:conf/iclr/JiangSQLZZY25,xiao2023chain,ahmaditeshnizi2024optimus,prasath2023synthesis,DBLP:journals/corr/abs-2303-08233}, where problem constraints are transformed into programs, enabling direct invocation of expert solvers to solve complex VRPs.
Its core value lies in democratizing advanced optimization tools, enabling non-expert users to bridge the gap between high-level business requirements and rigorous solver execution. Traditionally, converting real-world VRPs into solver-ready models requires substantial domain expertise, which limits the accessibility of powerful optimization engines such as Gurobi~\cite{gurobi2024} and OR-Tools~\cite{ortools}. The rapid development of LLMs has made this process increasingly feasible through automated reasoning and code generation. Existing automatic modeling methods mainly follow two routes: formulation-first methods, which first derive explicit mathematical formulations before program implementation, and code-only methods, which directly generate executable solver code from problem descriptions. In this paper, we use \emph{automatic modeling} as a broad term to refer to the process of transforming high-level problem descriptions into solver-ready optimization programs, regardless of whether an explicit mathematical formulation is generated as an intermediate step. 
% Recent state-of-the-art method DRoC~\cite{jiang2025droc} also adopt a code-only paradigm by leveraging retrieval-augmented generation (RAG) to incorporate solver documentation, which improves modeling accuracy for complex VRP variants. However, such approaches primarily rely on external knowledge and invoke LLMs for decision-making at each step, resulting in high inference latency while underutilizing accumulated experience from previous modeling attempts.
Recent state-of-the-art method DRoC~\cite{jiang2025droc} also adopts a code-only paradigm and improves the modeling accuracy of complex VRP variants by using retrieval-augmented generation (RAG). However, DRoC primarily depends on LLMs for action decision-making at each modeling step, leading to high inference latency caused by repeated LLM calls. In addition, its retrieval process mainly leverages external solver knowledge, overlooking valuable internal experience accumulated from previous modeling attempts. 
% In contrast, RLEA incorporates an evolving internal memory pool and trains a lightweight reinforcement-learning-based planner to make action decisions, thereby reducing reliance on costly LLM-based decision-making while better exploiting historical modeling experience.

In this work, we follow the automatic modeling paradigm and propose a Reinforcement Learning Enhanced LLM Agents famework (RLEA) that integrates a lightweight neural Planner, retrieval-based knowledge augmentation, and an evolved memory mechanism. This design enables efficient action orchestration while leveraging both external knowledge and internal experience to improve modeling robustness for complex VRP variants.
The contributions of this paper can be summarized as follow:
% % 目前的工作以及不足 本文的优势
% Existing literature on automatic modeling in operations research primarily focuses on general optimization problems~\cite{DBLP:conf/iclr/JiangSQLZZY25,xiao2023chain,ahmaditeshnizi2024optimus,prasath2023synthesis,DBLP:journals/corr/abs-2303-08233}. Prior studies, such as Chain-of-Expert~\cite{xiao2023chain} and OptiMUS~\cite{ahmaditeshnizi2024optimus}, propose multi-agent frameworks that coordinate multiple LLM agents through prompts. Despite performing well on general optimization tasks, these approaches often struggle to adapt to complex VRPs due to their reliance on designed task-specific prompts. DRoC~\cite{jiang2025droc} introduces a specialized framework that integrates Retrieval-Augmented Generation (RAG) to better handle complex VRPs. Following a similar paradigm to DRoC, our work bypasses the explicit output of intermediate mathematical formulations, directly transforming the problem description into executable solver code to streamline the modeling process. However, DRoC primarily relies on external knowledge, overlooking the critical role of internal experiential memory in refining problem-solving strategies. Furthermore, the reliance on LLM for every decision step often results in substantial latency.

% In this paper, we design a novel multi-agent framework that addresses the limitations mentioned above and makes the following distribution:
\begin{enumerate}
  \item We propose RLEA, a novel multi-agent framework that integrates reinforcement learning with LLM agents to automatically generate solver-ready programs for complex VRP variants.
  \item We introduce a lightweight Planner trained with Soft Q-learning to adaptively select actions for LLM agents, enabling efficient exploration while significantly reducing the latency compared with LLM-based decision-making.
  \item We present an agent equipped with evolvable memory. This agent actively analyzes interaction trajectories to identify successful action paths and derives the root causes of constraint conflicts from failure instances. Beyond the external knowledge provided by RAG, the agent further leverages its accumulated experience to enhance modeling robustness.
   \item We conducted a comprehensive evaluation of RLEA across 48 distinct VRP variants. 
   Experimental results demonstrate that RLEA significantly outperforms existing methods. Compared to the state-of-the-art (SOTA) method DRoC, RLEA achieves a 16.67\% increase in success rate and a 10.41\% reduction in runtime error rate.
   
\end{enumerate}

\section{Related Work}
% Existing research on leveraging Large Language Models (LLMs) to solve Operations Research (OR) problems can be broadly categorized into two paradigms based on the modeling depth and the workflow structure. The first paradigm focuses on a sequential process that translates natural language into formal mathematical symbols before implementation, while the second emphasizes a more direct mapping from problem descriptions to executable code.
Recent studies have explored the potential of LLMs for automating modeling of operations research problems. Existing approaches within this paradigm can be broadly categorized into two groups based on whether an explicit mathematical formulation is generated before program implementation.

% \subsubsection{Sequential Formalization and Implementation}
% Early works focused on establishing general frameworks for operations research problems. For instance, Xiao et al. propose a multi-agent framework composed of specific experts for automatically modeling and solving operations research problems~\cite{xiao2023chain}. AhmadiTeshnizi et al. present a simpler agent-based workflow to formulate and address mixed integer linear programming problems from their natural language definitions~\cite{ahmaditeshnizi2024optimus}. To enhance the generalization and accuracy of such sequential paradigms, subsequent research has focused on unified learning-based frameworks. A notable example is LLMOPT, which introduces a universal five-element formulation to model diverse optimization types, ranging from linear programming to complex combinatorial optimization. By employing multi-instruction tuning and self-correction mechanisms, LLMOPT improves the formalization process and mitigates hallucinations in code generation, achieving significant accuracy gains across various industrial fields~\cite{DBLP:conf/iclr/JiangSQLZZY25}. These methods follow a rigorous "Formalize-then-Implement" paradigm, prioritizing the clarity of mathematical symbolic logic before proceeding to code generation.

\subsubsection{Formulation-First Methods}
The first line of work follows a sequential pipeline that explicitly translates optimization problems into mathematical expressions before generating executable code. Early studies proposed structured workflows~\cite{ahmaditeshnizi2024optimus} to convert natural language descriptions into mathematical representations. Chain-of-Experts~\cite{xiao2023chain} introduces specialized agents to collaboratively construct mathematical formulations and corresponding solver implementations. 
More recently, unified learning-based frameworks have been proposed to improve the robustness of this formalization process. LLMOPT~\cite{DBLP:conf/iclr/JiangSQLZZY25} introduces a universal five-element formulation that enables LLMs to represent diverse optimization problems in a structured symbolic form before generating solver-ready programs. By emphasizing mathematical clarity and structured reasoning, these approaches enhance interpretability and reduce the risk of incorrect program implementations.
% \subsubsection{Direct to Implementation Modeling}
% While the sequential paradigm provides strong interpretability, the intermediate step of generating explicit mathematical formulations can be redundant and error-prone when dealing with highly specialized optimization tasks. Consequently, a more streamlined paradigm has emerged that bypasses formal mathematical output to directly map problem descriptions to executable code. To further address the complexity of specific domains like VRPs, Jiang et al. introduce a RAG approach~\cite{jiang2025droc}. By leveraging external solver documentation, their approach significantly improved the accuracy of generated programs for complex VRP variants. Notably, this "Direct-to-Implementation" strategy enhances modeling efficiency by reducing the cognitive load required for formula derivation. However, these frameworks rely heavily on LLMs for decision-making, which incurs prohibitive costs as API calls are required at every iteration.

\subsubsection{Code-Only Methods}
While methods that explicitly generate mathematical formulations offer strong interpretability, they often introduce unnecessary reasoning steps that can be error-prone, especially for highly specialized optimization tasks.
As a result, a more efficient paradigm has emerged that directly maps problem descriptions to executable solver code, bypassing the need for explicit mathematical formulation.
A representative example is DRoC~\cite{jiang2025droc}, which leverages RAG to incorporate solver documentation and constraint-specific knowledge during code generation. By grounding the generation process in external resources, DRoC significantly enhances the modeling accuracy of complex VRP variants.
However, such frameworks typically rely on LLMs for decision-making at every interaction step, resulting in high inference latency and underutilizing accumulated experience from previous modeling efforts.

To overcome these limitations, we propose RLEA, a reinforcement learning enhanced multi-agent framework~\cite{yang2026phase}. RLEA introduces a lightweight neural Planner that efficiently orchestrates LLM-driven actions. By integrating policy learning, retrieval-augmented knowledge, and an evolved memory mechanism, RLEA enables more efficient and robust automatic modeling for complex VRP variants.

\section{Preliminaries}
\subsection{Complex Vehicle Routing Problems}
The VRPs aim to determine the optimal routes for a fleet of vehicles serving a set of customers. Formally, the problem is represented on a directed graph $G = (V, E)$, where $V = {0, 1, \dots, N}$ denotes the set of nodes (node $0$ is the depot, and $V_c = \{1, \dots, N\}$ are the customers), and $E = \{(i, j) \mid i, j \in V, i \neq j\}$ defines the edges. Each edge $(i, j)$ has an associated travel cost $c_{ij}$, representing the distance between nodes $i$ and $j$. The binary decision variable $x_{ij}$ is 1 if a vehicle travels from node $i$ to $j$, and 0 otherwise. The objective is to minimize the total travel cost, expressed as:

\begin{equation}
\mathcal{J} = \min \sum_{i \in V} \sum_{j \in V, j \neq i} c_{ij} x_{ij}.
\end{equation}
This objective is subject to several constraints:

\begin{equation}
\label{degree_constraint}
\sum_{j \in V, j \neq i} x_{ij} = \sum_{j \in V, j \neq i} x_{ji} = 1, \quad \forall i \in V_c,
\end{equation}

\begin{equation}
\label{c2}
\sum_{j \in V_c} x_{0j} = \sum_{i \in V_c} x_{i0}.
\end{equation}

Eq.~\eqref{degree_constraint} ensures each customer is visited by exactly one vehicle, and Eq.~\eqref{c2} ensures the depot flow conservation, with the same number of vehicles returning to the depot.
We also consider nine additional VRP constraints based on real-world variants encountered in practical applications~\cite{jiang2025droc}: 1) Capacity, the vehicle load limit; 2) Open routes, where vehicles don’t have to return to the depot; 3) Distance limit, the total travel distance cannot exceed a specified bound; 4) Service time, each customer requires a specific service time; 5) Time window, each customer must be served within a given time frame; 6) Multiple depots, allowing vehicles to start and return to multiple depots; 7) Resource constraints, additional resource limitations; 8) Prize collecting, maximizing rewards while satisfying routing constraints; 9) Pickup and delivery, where pickup must occur before delivery. More details are in Appendix 1.

\subsection{Problem Formulation} %planner要先说一句！
%In RLEA, we introduce a Planner that learns to select actions to guide LLM agents in generating and refining executable modeling code for a given VRP variant. Subsequently, the generated code is executed to invoke existing expert solvers, obtaining the solution with the minimum objective value for the current problem instance. This process is modeled as a Markov Decision Process (MDP) augmented with a memory pool, defined by the tuple  $\langle \mathcal{S}, \mathcal{A}, \mathcal{P}, \mathcal{R}, \gamma, \mathcal{M} \rangle$. Each state $s \in \mathcal{S}$ consists of the embeddings of the problem description and the current modeling code. The action space $\mathcal{A}$ defines the operations that the agent can invoke to generate or revise the modeling code. %Upon executing an action, the environment transitions according to the state transition probability $\mathcal{P}$, resulting in a transition to a new state. 序列的意义
%To guide this sequence, The reward function $\mathcal{R}$ evaluates the quality and feasibility of the generated code, while the discount factor $\gamma \in [0, 1]$ balances immediate feedback against the importance of long-term optimization success. Additionally, a memory pool $\mathcal{M}$ stores historical experiences to facilitate learning from past successes and failures.

The goal of this work is to automatically generate solver-ready programs for complex VRP variants. Formally, We model this process as a Markov Decision Process (MDP) augmented with a memory pool, defined by the tuple
$\langle \mathcal{S}, \mathcal{A}, \mathcal{P}, \mathcal{R}, \gamma, \mathcal{M} \rangle$.

% In RLEA, a Planner that sequentially selects actions to guide LLM agents in generating and refining executable modeling code for a given VRP variant. Once the code is produced, it is executed to invoke existing expert solvers, and the best solution found for the current problem instance is returned.

At each step $t$, an agent follows a policy $\pi(a_t \mid s_t)$ to select an action $a_t \in \mathcal{A}$ based on the current state $s_t \in \mathcal{S}$. Repeated action selection induces a trajectory $\tau = (s_0, a_0, s_1, a_1, \dots, s_T)$, which represents the entire process of code generation and refinement. Each state consists of the embeddings of the problem description and the current modeling code, while the action space defines the operations that the agent can invoke to generate or revise the code. After executing an action, the environment transitions to a new state according to the transition probability $\mathcal{P}(s_{t+1} \mid s_t, a_t)$.
The reward function $\mathcal{R}$ evaluates the quality and feasibility of the generated code. In addition, a memory pool $\mathcal{M}$ stores historical experiences, allowing the agent to learn from past successes and failures.

% Importantly, our goal is to solve a code generation problem, rather than explicitly generating the mathematical formulation as in prior work.
% The memory pool$\mathcal{M}$ stores historical experiences, enabling the agent to refine its generated code based on past successes and failures in solving complex VRP variants.
To improve exploration and avoid premature convergence to sub-optimal actions, we adopt the Soft Q-Learning framework~\cite{haarnojaReinforcementLearningDeep2017}. This approach augments the standard reinforcement learning objective with an entropy regularization term, encouraging stochastic policies and better exploration. The resulting objective is defined as:
% \begin{equation}
% J(\pi) = \mathbb{E}_{\tau \sim \pi} \left[ \sum_{t=0}^{T} \gamma^t r(s_t, a_t) + \alpha \, {H}\!\left(\mu(\cdot \mid s_t, M_t)\right) \right]
% \label{eq:policy_objective}
% \end{equation}

\begin{equation}
J(\pi) = \mathbb{E}_{\tau \sim \pi} \left[ \sum_{t=0}^{T} \gamma^t r(s_t, a_t) + \alpha \, {H}\!\left(\pi(\cdot \mid s_t, M_t)\right) \right],
\label{eq:policy_objective}
\end{equation}
where $\tau$ denotes a trajectory of states, actions, and rewards generated by the policy, $H$ denotes the entropy, and $\alpha$ is a weighting coefficient that regulates the impact of the entropy. 
The discount factor $\gamma \in [0,1]$ balances immediate rewards with long-term optimization performance.

\section{Methodology}
\subsection{Overview}
% We propose an RL-assisted multi-agent framework that enables LLMs to adaptively select and execute the most effective actions tailored to the specific constraints and characteristics of the current problem. To this end, we design a multi-agent system consisting of a neural Planner, an LLM-based Executor, and a memory module. %By integrating internal and external knowledge sources with an autonomous self-correction mechanism, this framework significantly enhances the model's reasoning accuracy while reducing execution errors in complex optimization tasks. 
% This framework includes two phases: training and inference, as illustrated in Figure~\ref{figmethod}. The section first gives an introduction to the three designed actions, then specifically presents the training strategy of the Planner, and lastly demonstrates the inference procedure in the testing phase..

% We propose the Reinforcement Learning Enhanced LLM Agents framework (RLEA), which integrates a neural Planner to adaptively select actions for LLM agents, considering the specific constraints and characteristics of the problem at hand. RLEA is depicted in Figure~\ref{figmethod}. Specifically, we design a multi-agent system composed of a neural Planner, an LLM-based Executor, and a memory module.
% This section first introduces the three actions, then presents the training strategy of the Planner, and finally describes the inference procedure.

\begin{figure}[t] 
  \centering      
  \includegraphics[width=1\textwidth]{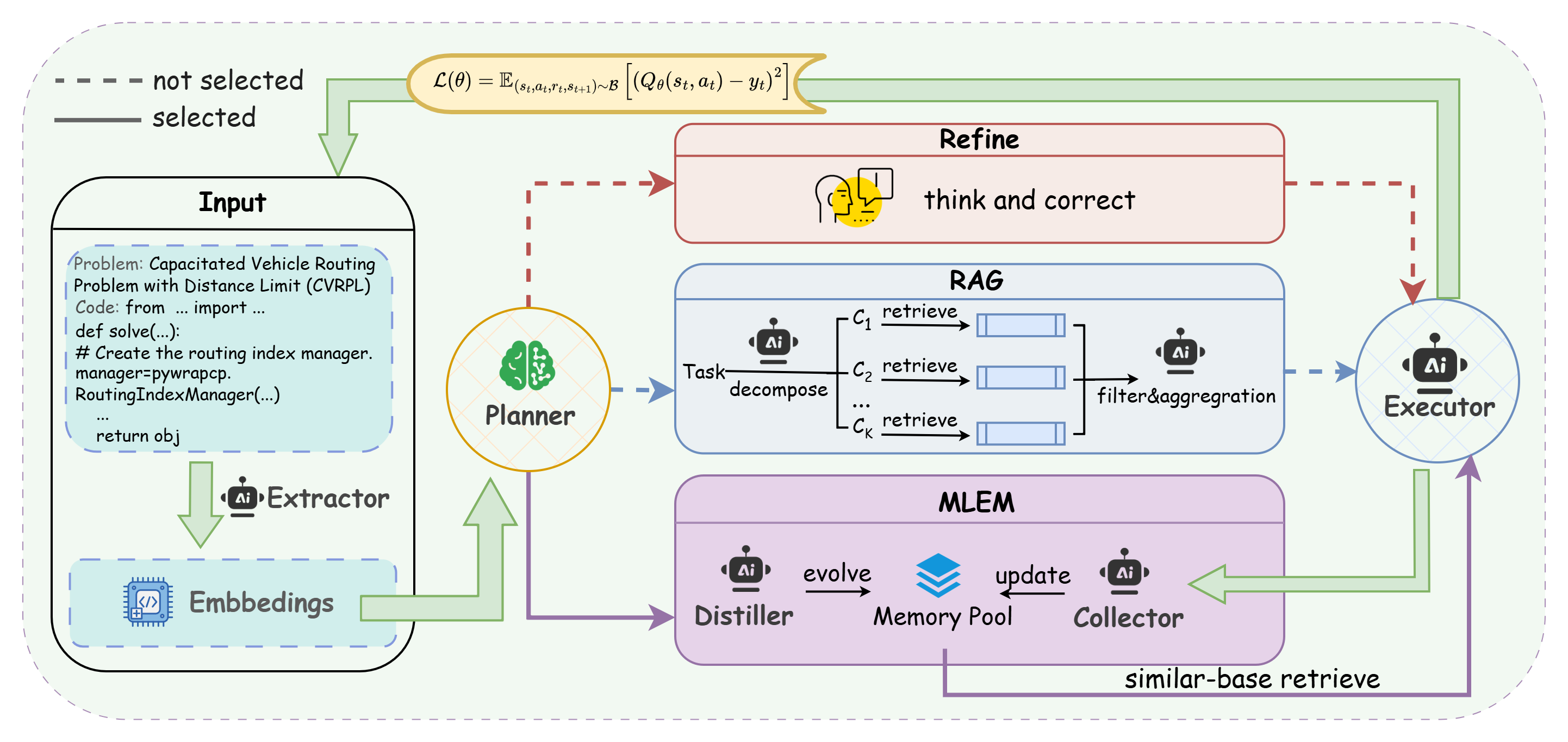} 
  \caption{Architecture of RLEA for automated VRP modeling. The input consists of the problem name and the current modeling code to be generated or revised. An extractor encodes this pair into embeddings, which define the Planner state. Based on the current state, the Planner follows a policy $\pi(a \mid s)$ to select the action, such as exploiting similar historical experiences from the memory pool (MLEM). The selected action is then executed to generate or update the modeling code and invoke the solver for the current problem instance. During training, the Planner samples actions according to the learned policy; during inference, the Planner is frozen and selects actions deterministically.}
  \label{figmethod}
\end{figure}

We propose Reinforcement Learning Enhanced LLM Agents (RLEA), a multi-agent framework for automatic VRP modeling. As illustrated in Figure~\ref{figmethod}, RLEA consists of three main components: a neural Planner, an LLM-based Executor, and a Memory module. The Planner selects actions based on the current problem state, the Executor carries out the selected action to generate or revise code, and the memory module continuously evolves historical interaction trajectories into reusable experience for future decision-making.

We next describe the action space, followed by the training strategy of the Planner and the inference procedure.

%Architecture of RLEA for Automated VRP Modeling. During the training phase, the input is first transformed into embeddings by the extractor. Based on these embeddings, the Planner samples actions according to a learned policy. The Executor then generates code based on the selected action and invokes the solver to solve the problem. In the inference phase, the Planner is frozen and deterministically selects actions.
\subsection{Execution Action Space Design}%To enable the agent to navigate complex problem-solving environments effectively,
%To enable agents to act efficiently in complex problem settings

To support automatic VRP modeling in complex settings, we design an action space $\mathcal{A}$ consisting of three actions: Refine, Retrieval-Augmented Generation (RAG), and Meta-Learning with Evolved Memory (MLEM). These actions are designed to address three key requirements in solver-oriented program generation: iterative error correction, the incorporation of external solver knowledge, and the reuse of evolved experience. Detailed prompts are provided in Appendix 2. Below we describe the mechanics of each action.

% Refine exploits the self-correction capability of LLMs to revise erroneous code, RAG introduces external solver knowledge to improve constraint modeling, and MLEM leverages distilled historical experiences to facilitate adaptation to new problem settings. 

\label{sec:action}
\subsubsection{Refine} 
% The Executor may generate erroneous code or hallucinations when handling complex VRP variants. However, LLMs can improve their output through iterative feedback and refinement, similar to human debugging. When this action is selected, the Executor checks whether all constraints are satisfied and analyzes possible directions for improving the current solution. It then revises the modeling code accordingly. This action allows the Executor to reflect on and logically correct its previous output. We denote the action as $a_r$.

When handling complex VRP variants, the executor may produce erroneous code or incomplete constraint implementations. The Refine action exploits the self-correction capability of LLMs by enabling iterative revision based on the current code state. When this action is selected, the executor examines whether the generated code satisfies all required constraints, identifies potential sources of failure, and revises the program accordingly. This process allows the executor to reflect on and correct its previous output in a debugging-style manner. We denote this action by $a_r$.

\subsubsection{Retrieval-Augmented Generation} 
To compensate for the limited solver-specific knowledge encoded in LLM parameters, we incorporate a RAG mechanism that injects external documentation into the code generation process. This action is denoted by $a_g$.

When selected, the Executor first decomposes the current problem $P$ into a set of atomic constraints,
\begin{equation}
\mathcal{C} = \{c_1, c_2, \dots, c_N\} = \Phi_{\text{decomp}}(P),
\end{equation}
where $\Phi_{\text{decomp}}$ denotes the LLM-based decomposition prompt and $N$ is the number of identified constraints, such as capacity limits or time windows. Each constraint $c_i$ is then used as a retrieval query.

Following~\cite{jiang2025droc}, both query keywords and external documents are encoded into embeddings. For each constraint $c_i$, we construct a query $Q_i$ using the template ``Python code for $c_i$''. The relevance between the query embedding $E(Q_i)$ and a document embedding $E(d_j)$ is measured by the squared Euclidean distance:
\begin{equation}
\mathcal{D}\big(E(Q_i), E(d_j)\big) = \|E(Q_i) - E(d_j)\|_2^2.
\end{equation}

The retrieved candidates are further filtered by the LLM to remove irrelevant documents. If multiple candidates remain, the LLM summarizes them and selects the most relevant reference. The Executor then generates or revises modeling code using the retrieved documentation associated with all constraints. If the RAG action is invoked again, the executor refines the failed code conditioned on the previously retrieved content.

\subsubsection{Meta-learning with Evolved Memory}
% By incorporating a instruction with the core idea of meta-learning, MLEM empowers the Executor to learn how to adapt to new problem settings rapidly. It serves as a dynamic knowledge repository that generalizes across tasks, allowing the model to extract high-level patterns from its memory to guide decision-making in unseen scenarios.The memory module consist of the Collector and the Memory Distiller. Formally, we define the memory pool as a set of tuples $\mathcal{M} = \{(q_i, f_i)\}_{i=1}^N$, where $q_i$ denotes the problem description and $f_i$ represents the execution feedback (success or failure analysis). The Collector monitors the Executor's generation process, summarizes the specific reasons attributing to the success or failure of each attempt and appends new tuples to $\mathcal{M}$.

To improve adaptation to new problem settings, we further introduce Meta-Learning with Evolved Memory (MLEM), which enables the agents to reuse evolved experience distilled from prior interactions. This action is denoted by $a_m$.

The memory module consists of a \textit{Collector} agent and a \textit{Memory Distiller} agent. Formally, the memory pool is defined as a set of tuples $\mathcal{M} = \{(q_i, f_i)\}_{i=1}^N$, where $q_i$ denotes the problem description and $f_i$ represents corresponding execution feedback, including success summaries or failure analyses. The Collector monitors the Executor's generation process, summarizes the causes of success or failure, and appends new records to the evolving memory pool $\mathcal{M}$.

% Upon activation, the Memory Distiller synthesizes raw records for each problem type. For a given problem $q$, it aggregates a subset $\mathcal{M}_q \subset \mathcal{M}$ into two types of experiences: the successful experience $f_{succ}$ and the failed experience $f_{fail}$.
% We then retrieve the top-$K$ most similar successful and failed memory pairs based on semantic similarity between problem embeddings:

When the MLEM action is activated for a current problem $q$, the Memory Distiller aggregates the subset $\mathcal{M}_q \subset \mathcal{M}$ associated with similar problem types and organizes it into two complementary forms of experience: successful experience $f_{\text{succ}}$ and failed experience $f_{\text{fail}}$. To retrieve the most relevant demonstrations, we rank historical cases by the semantic similarity between problem embeddings:

\begin{equation}
S(q_{curr}, q_i) = \frac{E(q_{curr}) \cdot E(q_i)}{\|E(q_{curr})\| \|E(q_i)\|},
\end{equation}
where $E(\cdot)$ denotes the embedding function. We then retrieve the top-$K$ most similar successful and failed memory pairs, denoted by $\mathcal{D}_{ret} = \{ (f_{succ}^{(k)}, f_{fail}^{(k)}) \}_{k=1}^K$.

The Executor implements the meta-learning mechanism by utilizing $\mathcal{D}_{ret}$ as in-context demonstrations. The action $a_m$ of MLEM is thus formalized as generating the solution conditioned on both the current problem $q_{curr}$ and the evolved memory ${D}_{ret}$. This mechanism enables fast adaptation to unseen constraints by leveraging historical meta-knowledge. The detailed prompt together with representative memory examples, are provided in Appendix 2.

\subsection{Training Phase: Policy Optimization and Memory Evolution}
Static LLM-based decision-making pipelines are often insufficient for complex VRP modeling~\cite{li2026phgpo}, where the most effective action depends on both the problem context and the current code state. This makes adaptive decision-making essential. In addition, the evolving memory pool is initially sparse, so the MLEM action $a_m$ may provide limited benefit in early training, which can further suppress exploration if a fixed action strategy is used. To address these challenges, RLEA employs a lightweight neural Planner trained with Soft Q-learning, enabling adaptive action selection without incurring the high cost of LLM-based decision-making at every interaction step.

The Planner learns a stochastic policy $\pi_\theta(a_t \mid s_t)$ over the action space $\mathcal{A}=\{a_r,a_g,a_m\}$, corresponding to Refine, RAG, and MLEM, respectively. The detailed definitions of these actions are given in Section~\ref{sec:action}. The policy is induced by a soft Q-network through the Boltzmann distribution:

% Static workflows lack the flexibility required to handle diverse VRP variants. Since the most effective strategy depends on the problem context and the current code state, the Planner must learn a stochastic policy rather than follow a fixed workflow. Meanwhile, the memory pool is empty at the beginning of training, causing the MLEM action $a_m$ to yield RLEAtively low rewards and discouraging early exploration. To address these challenges, we introduce a lightweight neural Planner optimized via Soft Q-learning, avoiding costly LLM-based decision-making at each step.

% The Planner learns a stochastic policy $\pi_\theta(a_t \mid s_t)$ over the action space $\mathcal{A}=\{a_r,a_g,a_m\}$, corresponding to Refine, RAG, and MLEM, respectively. Detailed descriptions of these actions are provided in Section~\ref{sec:action}. The policy is induced by the soft Q-network through a Boltzmann distribution:
\begin{equation}
    \pi_\theta(a \mid s) =
    \frac{\exp\left(Q_\theta(s,a)/\alpha\right)}
    {\sum_{a' \in \mathcal{A}} \exp\left(Q_\theta(s,a')/\alpha\right)},
\end{equation}
where $\alpha > 0$ is the temperature parameter controlling the exploration--exploitation trade-off. A larger $\alpha$ encourages broader exploration over candidate actions, which is particularly beneficial in the early stage when evolved memory is not yet sufficiently informative, whereas a smaller $\alpha$ leads to more greedy action selection.

During training, the planner interacts with the environment to generate a trajectory $\tau$. To guide policy optimization, we design a dense reward based on the optimality gap. Let $y_t^{\mathrm{obj}}$ denote the objective value of the solution obtained at step $t$. The relative optimality gap $\delta_t$ is defined as
\begin{equation}
    \delta_t = \left| \frac{y_t^{\mathrm{obj}} - y^*}{y^*} \right|,
\end{equation}
where $y^*$ denotes the reference optimal objective value obtained from expert benchmarks.
To encourage high-quality solutions while explicitly penalizing execution failures, the reward $r_t$ is defined as
\begin{equation}
    r_t =
    \begin{cases}
    \frac{1}{1+\delta_t}, & \text{if a feasible solution } y_t^{\mathrm{obj}} \text{ is obtained}, \\[4pt]
    0, & \text{otherwise}.
    \end{cases}
\end{equation}
This reward is bounded in $(0,1]$ for valid solutions and approaches 1 as the solution converges to the optimum.

% The Planner is parameterized by a soft Q-network $Q_\theta(s,a)$. Its input state consists of the embedding of the problem context, including the VRP variant description and the current modeling code, produced by a frozen open-source small language model serving as the extractor. The network is trained by minimizing the Soft Bellman residual. Specifically, the parameters are updated toward the soft temporal-difference target:
% \begin{equation}
%     \hat{y}_t = r_t + \gamma (1-d_t)\,\alpha \log \sum_{a' \in \mathcal{A}}
%     \exp\left(\frac{Q_{\bar{\theta}}(s_{t+1},a')}{\alpha}\right),
% \end{equation}
% where $d_t \in \{0,1\}$ is the termination flag and $Q_{\bar{\theta}}$ denotes the target network. The training loss is defined as:
% \begin{equation}
%     \mathcal{L}(\theta) =
%     \mathbb{E}_{(s_t,a_t,r_t,s_{t+1}) \sim \mathcal{B}}
%     \left[
%     \left(Q_\theta(s_t,a_t)-\hat{y}_t\right)^2
%     \right],
% \end{equation}
% where $\mathcal{B}$ is the experience replay buffer. Note that $\mathcal{B}$ is used for RL optimization, whereas the memory pool $\mathcal{M}$ is maintained separately for the MLEM action to retrieve and evolve useful historical cases.

The Planner is parameterized by a soft Q-network $Q_\theta(s,a)$. Its input state is the embedding of the current problem context, including the VRP variant description and the current modeling code, extracted by a frozen open-source small language model. The Planner is trained by minimizing the soft Bellman residual. Specifically, the soft temporal-difference target is defined as
\begin{equation}
    \hat{y}_t = r_t + \gamma (1-d_t)\,\alpha \log \sum_{a' \in \mathcal{A}}
    \exp\left(\frac{Q_{\bar{\theta}}(s_{t+1},a')}{\alpha}\right),
\end{equation}
where $d_t \in \{0,1\}$ is the termination flag, $\gamma \in [0,1]$ is the discount factor, and $Q_{\bar{\theta}}$ denotes the target network. The training objective is
\begin{equation}
    \mathcal{L}(\theta) =
    \mathbb{E}_{(s_t,a_t,r_t,s_{t+1}) \sim \mathcal{B}}
    \left[
    \left(Q_\theta(s_t,a_t)-\hat{y}_t\right)^2
    \right],
\end{equation}
where $\mathcal{B}$ is the experience replay buffer.
Meanwhile, memory evolution proceeds in parallel with policy optimization. The replay buffer $\mathcal{B}$ is used exclusively for reinforcement learning updates, whereas the memory pool $\mathcal{M}$ is maintained separately to accumulate and evolve historical interaction records for the MLEM action. 

% This separation ensures that policy learning focuses on action-value estimation, while memory evolution continually improves the quality of the evolved experience available for future retrieval.

\subsection{Inference Phase: Policy Evaluation and Generalization}
% In the inference phase, the parameters $\theta$ of the Planner are frozen. The Executor first attempts to directly generate a solution for the given problem. If this attempt fails, the problem description together with the generated code forms the initial state for the Planner.
% The Planner then performs a forward pass and selects actions from the action space $\mathcal{A}$ according to the learned policy. Guided by the selected actions, the Executor iteratively generates or refines the modeling code for the target VRP variant.

During inference, the Planner parameters $\theta$ are frozen. Given a target VRP variant, the executor first attempts to generate solver-ready code from the problem description directly. If this initial attempt fails to produce a valid code, the problem description together with the generated code is encoded as the initial state for the Planner.

Starting from this state, the Planner performs a forward pass and selects actions from the action space $\mathcal{A}$ according to the learned policy. Conditioned on the selected actions, the Executor iteratively generates or refines the modeling code for the target problem. In this way, the Planner serves as a lightweight decision-making module that orchestrates the interaction process without requiring costly LLM-based deliberation at every step.

We impose a maximum number of steps $T_{max}$. The procedure terminates early once the generated code successfully invokes the solver and produces a solution whose relative optimality gap is below 5\%.
This mechanism prevents infinite refinement loops while still allowing sufficient opportunities for the agent to correct modeling or syntax errors.

\section{Experiments}
% In this section, we conduct comprehensive experiments to verify the performance of our approach. All the experiments are conducted on a Tesla A40 GPU and Intel(R) Core(TM)i5-7500 CPU. We compare our approach with other baselines on 48 VRP variants composed of different constraints. In our experiments, we employ Qwen(Qwen2.5-1.5B-Instruct) as the Extractor, ChatGPT(gpt-4o-2024-08-06) as the Memory Agents and DeepSeek(deepseek-reasoner) as the Executor. For offline inference, the Executor performs predictions with the pre-trained memory pool. For online inference, instead of relying on pre-trained memory, the Memory Agent dynamically populates and updates the memory pool in a test-time learning fashion. We primarily validate RLEA on two widely-used solvers, Gurobi and OR-tools. The number of attempts in inference is set as $T_{max} = 6$. To ensure a fair comparison, we use the same times of iterations across all baselines. The experimental results are the average of three independent experiments. 

%说一下图4
In this section, we conduct comprehensive experiments to evaluate the effectiveness of RLEA on automated VRP modeling. The experiments cover 48 VRP variants composed of different constraint combinations. All experiments are conducted on a Tesla A40 GPU and an Intel i5-7500 CPU.

% In our framework, Qwen2.5-1.5B-Instruct is used as the Extractor, DeepSeek-Reasoner as the Executor, and ChatGPT (gpt-4o-2024-08-06) as the Memory Agent. We evaluate two inference settings. In \textit{offline inference}, the Executor generates solutions using a pre-trained memory pool. In \textit{online inference}, the Memory Agent dynamically constructs and updates the memory pool during test time.

% We evaluate RLEA on two widely-used optimization solvers, Gurobi and OR-Tools. The maximum number of interaction steps during inference is set to $T_{max}=6$. To ensure a fair comparison, all baselines use the same iteration budget. Reported results are averaged over three independent runs. All experiments are conducted on a Tesla A40 GPU and an Intel i5-7500 CPU.

\subsection{Experiment Settings}
\subsubsection{Hyperparameters}
% Throughout all experiments, we train the Planner using the Soft Q-Learning algorithm. The Q-network parameters are updated using the Adam optimizer with a learning rate of $1 \times 10^{-4}$. The architecture consists of a pre-trained LLM backbone and a trainable MLP head. We employ Qwen2.5-1.5B-Instruct as the backbone to encode state representations, with a maximum token length of 8192. The backbone parameters are frozen during training to ensure computational efficiency. The trainable Q-value head is a neural network comprised of one hidden layer with 256 hidden units and ReLU nonlinearity, mapping the state embeddings to the action space.

% The algorithm maintains a replay memory of size 50,000. Training commences once the replay pool contains at least 3 samples. For each learning step, a mini-batch of size 1 is sampled uniformly. The discount factor $\gamma$ is set to 0.99. To balance exploration and exploitation, the entropy temperature coefficient $\alpha$ is fixed at 4. We perform a hard update on the target network parameters every 4 learning steps by directly copying weights from the online network. Each training episode is limited to a maximum of 16 time steps to prevent infinite loops in the interaction process.
The Planner is trained using the Soft Q-Learning algorithm with the Adam optimizer and a learning rate of $1\times10^{-4}$. The model consists of a frozen small language model (SLM) and a trainable Q-value head. We adopt Qwen2.5-1.5B-Instruct as the SLM to encode state representations with a maximum context length of 8192 tokens. The Q-value head is a two-layer MLP with 256 hidden units and ReLU activation, mapping state embeddings to the action space. DeepSeek-Reasoner as the Executor and ChatGPT (gpt-4o-2024-08-06) as the Memory Agent.

The replay buffer has a capacity of 512 state transitions, with a training batch size of 16. The discount factor is set to $\gamma=0.99$, and the entropy temperature coefficient is fixed at $\alpha=4$ to encourage exploration. The target network is updated every four training steps via hard parameter copying.
Each training episode is limited to a maximum of 16 interaction steps to avoid infinite loops.

\subsubsection{Baselines}
We compare RLEA with five representative baselines:
\begin{itemize}
    \item \textbf{Direct Generation:} Standard prompting without additional reasoning or correction mechanisms.
    \item \textbf{Reasoning-based methods:} Self-Refine~\cite{madaan2023self} and Chain-of-Thought (CoT)~\cite{zhang2024solving}.
    \item \textbf{Formulation-first approach:} Chain-of-Experts (CoE)~\cite{xiao2023chain}.
    \item \textbf{Code-only approach:} DRoC~\cite{jiang2025droc}.
\end{itemize}

In our framework, two inference settings are evaluated. In \textit{offline inference}, the Executor generates solutions using a pre-trained memory pool. In \textit{online inference}, the memory agent dynamically constructs and updates the memory pool during test time.

% We evaluate RLEA on two widely-used optimization solvers, Gurobi and OR-Tools. The maximum number of interaction steps during inference is set to $T_{max}=6$. To ensure a fair comparison, all baselines use the same iteration budget. Reported results are averaged over three independent runs. All experiments are conducted on a Tesla A40 GPU and an Intel i5-7500 CPU.
To ensure robustness, we conducted evaluations using two widely used optimization solvers: Gurobi~\cite{gurobi2024} and OR-Tools~\cite{ortools}. The maximum number of interaction steps during inference was set to $T_{max}=6$. To ensure a fair comparison, all baselines were assigned the same iteration budget. As in previous work~\cite{jiang2025droc}, the reported results are the average of three independent runs.

\subsubsection{Performance Metrics}
There are two performance metrics we used:
\begin{itemize}
\item \textbf{Success Rate (SR):} This metric evaluates the capability of the framework to generate valid modeling code. It is defined as:
\begin{equation}
\text{SR} = \frac{N_{succ}}{N_{total}} \times 100\%
\end{equation}
where $N_{total}$ is the total number of VRP instances tested. $N_{succ}$ represents the number of instances where the generated code successfully executes and yields a feasible solution (i.e., the solver successfully returns a valid objective value without errors).
\item \textbf{Runtime Error Rate (RER):} This metric reflects the proportion of programs that fail to execute due to syntax errors, API misuse, or internal logical flaws. It is calculated as:
\begin{equation}
    \text{RER} = \frac{N_{err}}{N_{total}} \times 100\%
\end{equation}
\end{itemize}

\subsection{Main Results}
Table~\ref{tab:performance_comparison} reports the performance of RLEA and five representative baselines in terms of Success Rate (SR) and Runtime Error Rate (RER) on both OR-Tools and Gurobi solvers.

Methods that rely solely on the internal knowledge of LLMs, including Standard Prompting, CoT, and Self-Refine, achieve limited performance, indicating that complex VRP variants are difficult to model accurately without domain-specific knowledge. CoE produces the lowest rates on SR due to its focus on Mixed-Integer Programming rather than complex VRPs, it maintains a remarkably low RER suggesting the potential of the multi-agent framework. By incorporating external solver documentation, DRoC improves performance over these methods, demonstrating the importance of domain-specific knowledge retrieval.

Our method achieves the best overall performance. On OR-Tools, RLEA reaches an SR of 62.50\%, outperforming the previous state-of-the-art DRoC by 16.67\%, while also significantly reducing runtime errors. Although the generated solutions are not always optimal, they already provide valid solver implementations, allowing human developers to focus on solution improvement rather than constructing models from scratch.
This improvement stems from the synergy of the learned Planner policy, external knowledge retrieval, evolved memory, and the refine mechanism, which together enable effective error correction and solution refinement.

In the online inference setting, the performance is slightly lower than in the offline setting but still surpasses all baselines. This difference is mainly due to the dynamic updating of the memory pool during inference, which introduces additional stochasticity in the learning process. However, this dynamic memory mechanism encourages broader exploration and ultimately contributes to a further reduction in runtime errors.

\begin{table}[t]
\caption{Performance comparison of different prompting and agent-based methods.}
\label{tab:performance_comparison}
\centering
\small
\begin{tabular}{l cc cc}
\hline
\multirow{2}{*}{Methods}
& \multicolumn{2}{c}{OR-Tools}
& \multicolumn{2}{c}{Gurobi} \\
\cline{2-5}
& SR (\%) & RER (\%) & SR (\%) & RER (\%) \\
\hline
Standard Prompting   & 29.17 & 39.58 & 29.17    & 27.08    \\
Self-Refine          & 31.25    & 39.58    & 33.33    & 35.42    \\
Chain of Thoughts    & 25.00    & 54.17    & 18.75    & 56.25    \\
Chain of Experts     & 16.67    & 27.08    & 8.33    & 18.75    \\
DRoC                 & 45.83 & 27.08 & 35.42 & 33.33 \\
\hline
Ours (offline)       & \textbf{62.50} & 16.67 & \textbf{43.75} & \textbf{16.67} \\
Ours (online)        & 60.42 & \textbf{12.50} & 39.58 & 22.92 \\
\hline
\end{tabular}
\end{table}

\subsection{Ablation Study}

\begin{table}[t]
\centering
\caption{Ablation study on the Planner.}
\label{tab:ablationplanner}
\small
\begin{tabular}{lccc}
\toprule
\textbf{Method} & SR (\%) & RER (\%) & Avg Time (s) \\
\midrule
\textbf{Ours} & \textbf{62.50} & 16.67 & \textbf{3.06} \\
w/o Planner & 50.00 & 14.58 & - \\
Prompt-based Planner & 56.25 & \textbf{12.50} & 153.30 \\
\bottomrule
\end{tabular}
\end{table}
We conduct ablation studies to evaluate the contribution of key components in RLEA, including the neural Planner and the action modules. The results are presented in Table~\ref{tab:ablationplanner} and Table~\ref{tab:ablation_study_ortools}.

\subsubsection{Ablation on the Planner} 
% To access the effect of the pre-trained Planner, we conduct the experiments on OR-Tools. For the baseline without a Planner, actions are selected via a random selection. In the Prompt-based variant, we employ the DeepSeek-Reasoner as the core decision-making engine to generate execution policies. As shown in Table~\ref{tab:ablationplanner}, RLEA significantly outperforms both baselines in terms of accuracy, attributable to its ability to construct coherent logical chains unlike the random baseline. The prompt-based Planner also achieves a respectable success rate, demonstrating that large language models can indeed formulate effective strategies by utilizing their vast internal knowledge and inherent reasoning capabilities. However, the inference latency is RLEAtively higher than RLEA, while our pre-trained neural Planner provides rapid decision-making. Regarding the Running Error Rate (RER), although RLEA shows a marginal increase compared to the baselines, this reflects a fundamental trade-off between high-performance planning and execution risk. By actively exploring more sophisticated and long-horizon logical chains to push the upper bound of task completion, our Planner inevitably encounters complex edge cases. In contrast, the lower RER of the baselines stems from their simpler or more cautious trajectories, which avoid such boundary risks but fail to achieve high accuracy.
To assess the impact of the learned Planner, we compare RLEA with two variants on OR-Tools: (1) w/o Planner, where actions are randomly selected, and (2) prompt-based Planner, where DeepSeek-Reasoner generates the execution strategy through prompting. As shown in Table~\ref{tab:ablationplanner}, RLEA achieves the highest SR, indicating that the learned policy effectively captures action-selection patterns across different VRP variants. Although the prompt-based Planner also attains a reasonable success rate, it incurs higher inference latency, while our neural Planner enables significantly faster decision-making. The slightly higher RER of RLEA reflects a trade-off between aggressive exploration and execution stability: by exploring more complex reasoning trajectories, the Planner occasionally encounters challenging edge cases. In contrast, the lower RER of the baselines stems from their simpler or more cautious trajectories, which avoid such boundary risks but fail to achieve high SR.
\begin{figure}[t] 
  \centering      
  \includegraphics[width=0.38\textwidth]{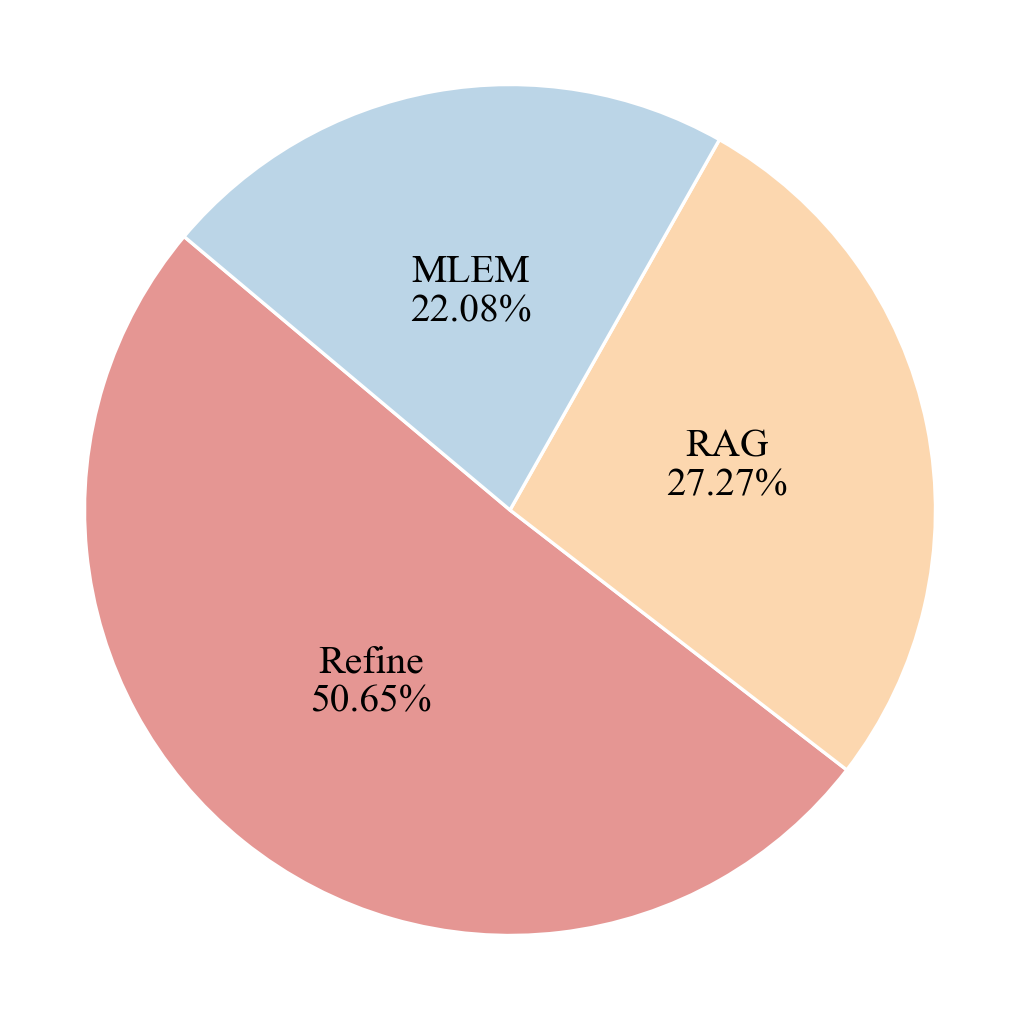} 
  \caption{Distribution of action types executed by RLEA on OR-Tools.} 
  \label{figaction}
\end{figure}
\subsubsection{Ablation on action module} In order to investigate the impact of the actions, we conduct an ablation study as shown in Table~\ref{tab:ablation_study_ortools}. The results demonstrate that our full method consistently achieves the highest SR and the lowest RER compared to all variants lacking a specific action module. 
Removing Refine causes the largest performance drop, suggesting that refine is the most critical action in our framework. Given that many modeling or programming errors are relatively easy to solve, the Executor can rectify minor logical flaws autonomously, thereby preventing simple errors from escalating into task failures. The action distribution in Figure~\ref{figaction} further supports this observation, where Refine accounts for 50.65\% of the executed actions. In contrast, removing RAG or MLEM leads to comparable performance degradation, suggesting that external knowledge retrieval and evolved memory play complementary roles in improving modeling accuracy.

\begin{table}[!htbp]
\caption{Ablation study of different actions on OR-Tools.}
\label{tab:ablation_study_ortools}
\centering
\small
\begin{tabular}{l cc}
\hline
Method & SR (\%) & RER (\%) \\
\hline
\textbf{Ours} & \textbf{62.50} & \textbf{16.67} \\
\hline
w/o Refine & 39.58 & 36.25 \\
w/o RAG & 43.75 & 22.92 \\
w/o Meta learning with evolved memory & 41.67 & 27.08 \\
\hline
\end{tabular}
\end{table}

\subsection{Impact of Different LLM Compositions}
The experimental results in Table \ref{tab:memory_action} reveal a clear performance advantage in heterogeneous multi-agent collaboration, specifically validating the optimal division of Action (reasoning-intensive models) and Memory (general-purpose models). The DeepSeek-r1 (Action) and GPT-4o (Memory) configuration achieves a peak Success Rate (SR) of 62.50\%, demonstrating that DeepSeek-r1 demonstrates its ability to leverage reasoning capabilities for the effective modeling of complex VRPs, while GPT-4o provides a robust contextual foundation for retrieval.

Conversely, when assigning GPT-4o to execute action and DeepSeek-r1 to manage the memory modules, the SR to drop to 45.83\%. This suggests that DeepSeek-r1’s intensive reasoning may be counterproductive for memory distillation, where its tendency toward over-deliberation introduces noise that hinders concise information flow. Furthermore, all single-model setups exhibit significant performance bottlenecks, such as the standalone GPT-4o’s low 37.50\% SR. These findings underscore that decoupling roles within a multi-agent framework prevents the cognitive overload inherent in single-agent architectures, allowing specialized LLMs to leverage their distinct strengths for superior collective performance.
\begin{table}[htbp]
\centering
\caption{Performance comparison of different LLM compositions}
\label{tab:memory_action}
\small
\setlength{\tabcolsep}{6pt}
\begin{tabular}{c|cccccc}
\toprule
\multirow{2}{*}{\diagbox[width=8em, height=2.5em]{\textbf{Executor}}{\textbf{Memory}}} 
& \multicolumn{2}{c}{DeepSeek-r1} 
& \multicolumn{2}{c}{GPT-4o} 
& \multicolumn{2}{c}{Qwen3-max} \\
\cmidrule{2-7}
& SR(\%) & RER(\%)
& SR(\%) & RER(\%) 
& SR(\%) & RER(\%) \\
\midrule
DeepSeek-r1 
& 52.08 & 22.92 
& \textbf{62.50} & 16.67 
& 43.75 & \textbf{12.50} \\

GPT-4o      
& 45.83 & 25.00 
& 37.50 & 31.25 
& 47.92 & 18.75 \\

Qwen3-max   
& 52.08 & 16.67 
& 60.42 & \textbf{12.50} 
& 45.83 & 14.58 \\
\bottomrule
\end{tabular}
\end{table}

\begin{figure}[htbp] 
  \centering      
  \includegraphics[width=0.38\textwidth]{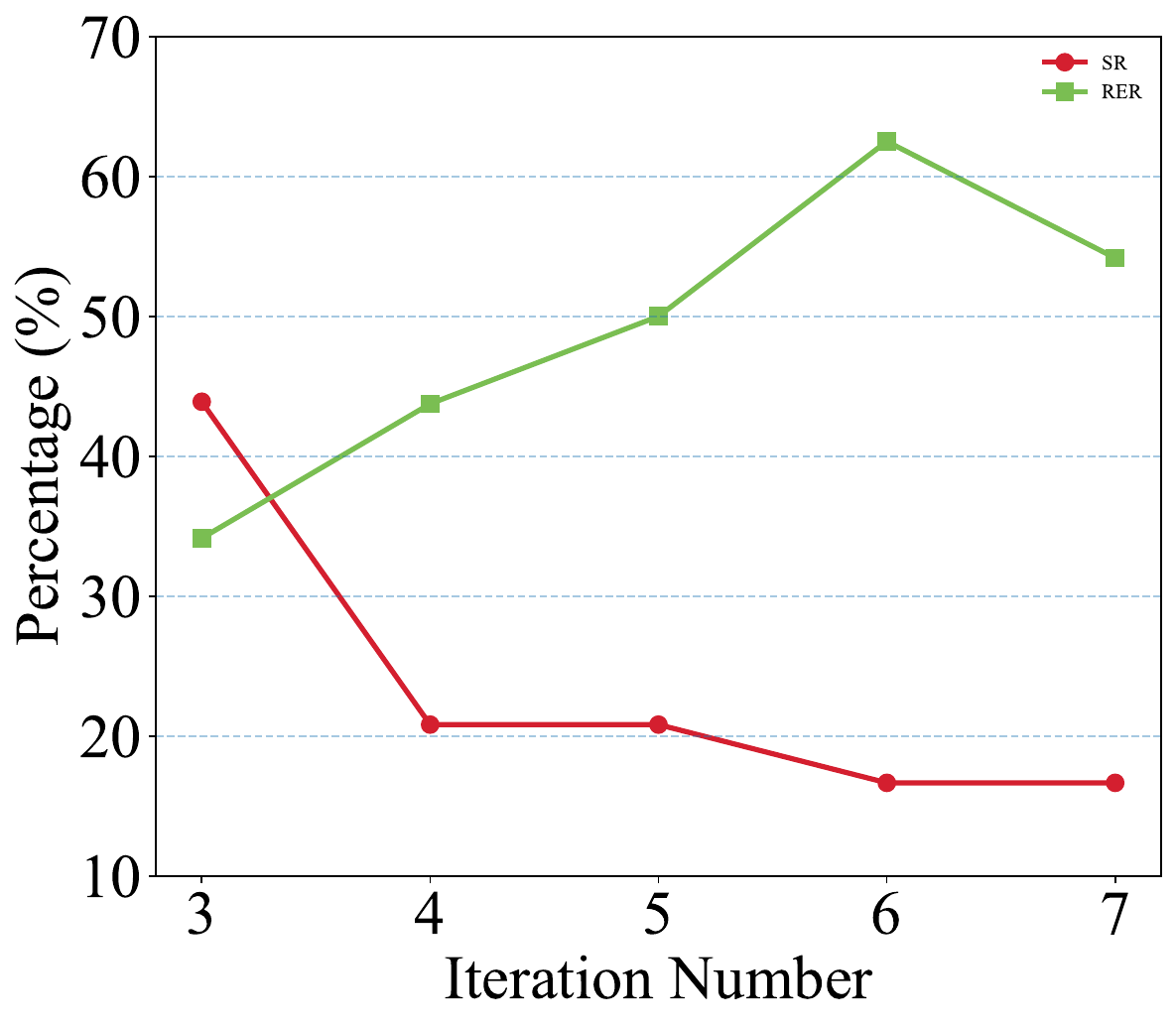} 
  \caption{Sensitivity analysis of OR-Tools performance relative to iterations} 
  \label{figiter}
\end{figure}
\subsection{Sensitivity Analysis of Iterations}
% We evaluate the impact of the number of iterations on OR-Tools. Since there are three actions in the action space in total, we set the minimal number of iterations as 3. From the results shown in Figure~\ref{figiter}, we can see there is a significant positive corRLEAtion between the number of iterations and performance. As the iterations increase from 3 to 5, the SR improves steadily from 34.15\% to 50.00\%, while the RER drops sharply from 43.90\% to 20.83\%. When the number of iterations reaches 6, the SR and RER both achieve the best results. Increasing the iterations beyond 6 leads to diminishing returns. This suggests that excessive iterations may lead to over-searching or algorithmic stagnation, where the added computational cost no longer yields better accuracy. Therefore, we set the number of iterations as 6 across our main experiments.

We analyze the effect of the maximum number of interaction iterations on performance using OR-Tools. Since the action space contains three actions, the minimum number of iterations is set to 3. As shown in Figure~\ref{figiter}, performance improves steadily as the number of iterations increases from 3 to 6: the SR rises from 34.15\% to 62.50\%, while the RER decreases from 43.90\% to 16.67\%.

When the iteration number reaches 6, the framework achieves its best performance. Further increasing the iteration budget brings only marginal gains. This suggests that excessive iterations may lead to over-searching or algorithmic stagnation, where the added computational cost no longer yields better performance. 
\section{Conclusion}
%In this paper, we propose RLEA, a novel multi-agent framework to automate the modeling and solving of complex VRPs. By integrating a lightweight Planner optimized through Soft Q-Learning, RLEA can rapidly conduct the Executor to execute actions, which reduces the computational overhead. Furthermore, the hybrid memory mechanism—combining evolutionary trajectory analysis with external RAG-based knowledge—provides a robust foundation for self-correction, enabling the agent to resolve intricate constraint conflicts that typically cause solver failures.The empirical results confirm that this architecture is highly effective, yielding a 62.50\% success rate across a rigorous benchmark of 48 diverse VRP variants, which represents a substantial 16.67\% performance gain over the previous state-of-the-art.

%In the future, we will explore the integration of multi-modal inputs to handle VRPs described through visual diagrams and the extension of this framework to dynamic, real-time routing scenarios where constraints evolve during execution.
In this paper, we propose RLEA, a novel multi-agent framework for automating the modeling and solving of complex VRPs. By integrating a lightweight Planner optimized via Soft Q-Learning, RLEA enables rapid execution by the Executor, reducing computational overhead. The hybrid memory mechanism, combining evolutionary trajectory analysis with external RAG-based knowledge, ensures robust self-correction, helping resolve constraint conflicts that often cause solver failures. Empirical results show a 62.50\% success rate across 48 VRP variants, achieving a 16.67\% performance gain over the previous state-of-the-art. Future work will explore integrating multi-modal inputs for VRPs described through visual diagrams and extending the framework to dynamic, real-time routing scenarios with evolving constraints.

\clearpage
\section*{Acknowledgements}
This work is supported by the National Natural Science Foundation
of China (62472461), and the Guangdong Basic and Applied Basic Research Foundation (2025A1515010129).
% \input{appendix}

\bibliographystyle{splncs04}
\bibliography{mybibliography}

\end{document}